\documentclass[10pt,twocolumn,letterpaper]{article}

\usepackage{cvpr}
\usepackage{times}
\usepackage{epsfig}
\usepackage{graphicx}
\usepackage{amsmath}
\usepackage{amssymb}
\usepackage{xcolor}

\usepackage{caption}
\usepackage{subcaption}
\usepackage[breaklinks=true,bookmarks=false]{hyperref}

\cvprfinalcopy 

\def\cvprPaperID{****} 

\ifcvprfinal\fi
\begin{document}

\title{Dealing with sequences in the RGBDT space}

\author{Gabriel Moy\`a\\
{\tt\small gabriel.moya@uib.es}
\and
Antoni Jaume-i-Cap\'o\\
{\tt\small antoni.jaume@uib.es}
\and
Javier Varona \\
{\tt\small xavi.varona@uib.es}
\and
Departament de Ci\`{e}ncies Matem\`{a}tiques i Inform\`{a}tica \\
Universitat de les Illes Balears, Cra. de Valldemossa, km 7.5. Palma, E-07122, Spain}

\maketitle

\begin{abstract}
 Most of the current research in computer vision is focused on working with single images without taking in account temporal information. We present a probabilistic non-parametric model that mixes multiple information cues from devices to segment regions that contain moving objects in image sequences. We prepared an experimental setup to show the importance of using previous information for obtaining an accurate segmentation result, using a novel dataset that provides sequences in the RGBDT space. We label the detected regions ts with a state-of-the-art human detector. Each one of the detected regions is at least marked as human once.
\end{abstract}

\section{Introduction}
\label{sec:Introduction}

Deep learning approaches have proven to solve per frame problems accurately. Most of the current research in computer vision is focused on working with single images without taking in account temporal information. Our point of view is that the key for solving a wide range of computer vision problems rely on the concept of the sequence. In a sequence, part of the necessary information to solve the problem on the current frame is given by the previous ones.

Tasks that are considered simple like detecting changes in a scene and segmenting the foreground from background, or detecting moving objects in image sequences are still challenging.

\subsection{The RGBDT space}

Color information is the most used feature in the computer vision history, standard cameras only capture this type of information. In~\cite{Toyama1999} several important challenges  color information were described, such as: shadows, changes in scene illumination,  camouflage and foreground aperture. Classical problems based on color information continue to be challenging for modern approaches, as described in~\cite{Sobral2014}, where 29 different algorithms were evaluated and compared. A feasible solution to overcome the limitations of the classical color-based approaches consists of adding new information to our proposed algorithms. Nowadays we can find devices that provides us novelty cues like depth or thermal information.

Depth sensors provide geometrical information about the scene where each pixel value represents the distance from the device to the point in the real world. The depth channel differs in its characteristics from color channels. In particular, it has a significant amount of missing information from instances in which the sensor is unable to obtain information at certain pixels. Depth devices suffer several problems such as: depth camouflage, specular materials, near objects, remote parts of the scene, non reachable areas and shadows which were described in \cite{moya2017modeling} .

Thermal imagery comes from passive sensors that capture the infrared radiation emitted by all objects with a temperature range, so instead of color or geometry it adds temperature information and eliminates the typical illumination problems of normal greyscale and RGB cameras. Some problems of this type of information are: reflections of the thermal radiation and a halo effect can also be observed
around warm items~\cite{gade2014thermal}.

\subsection{Background subtraction}

There is a large body of literature on the subject of background subtraction, here we focus on approaches that fuse more than on information cue.  Most of the explained techniques modify traditional background subtraction approaches by applying the same algorithm for depth  or thermal (in addition to the color channels) and suggesting some heuristics to address the heterogeneous characteristics of these different cues.

Harville~\textit{et al.} presented an approximation to Gaussian mixture modeling at each pixel~\cite{Harville2001}. A multidimensional Gaussian mixture distribution was constructed, with three components in a luminance-normalized color space and one depth channel. Special processing was performed to address absent depth pixels. No update phase was described; therefore, this algorithm can only be used in static scenes. 

Hofmann and Rigoll also proposed a Mixture of Gaussians approach  where depth and infrared data are combined to detect foreground objects. Each pixel was classified by binary combinations of foreground masks. The performance of this approach is limited because a failure of one of the models affects the final pixel classification~\cite{Hofmann}.

Camplani~\textit{et al}. proposed a \textit{per-pixel} background modeling approach that fuses different statistical classifiers based on depth and color data by means of a weighted average combination~\cite{Camplani2014}. A mixture of Gaussian distributions was used to model the background, and a uniform distribution was used for modeling the foreground. Same authors presented another approach in~\cite{Camplani2014a} based on the fusion of multiple region-based classifiers. Foreground objects were detected by combining a region-based foreground depth data prediction with different background models. The information given by these modules is fused in a mixture-of-experts fashion to improve the foreground detection accuracy.

Leens~\textit{et al}. presented a new approach using RGB and ToF (\textit{Time-of-Flight}) cameras based on a Parzen windows-like process. Each model was processed independently and the foreground masks are then combined using logical operations and then post-processed with morphological operators~\cite{JeromeLeensSebastienPierardOlivierBarnich1978}. 

Enrique Fernandez-Sanchez~\textit{et al.}~proposed an adaptation of the Codebook~\cite{Kim2004} background subtraction algorithm using a four-channel codebook~\cite{Fernandez-Sanchez2013}. Depth information was also used to bias the distance in chromaticity space associated with a pixel according to the depth measurements. Therefore, when the depth value is invalid, the detection depends entirely on color information.

Clapes \textit{et al}. presented a background subtraction technique in which a four-dimensional Gaussian distribution was used as the first step of the user identification and object recognition surveillance system. As they used a single Gaussian approximation, the algorithm was not able to manage multi-modal backgrounds~\cite{Clapes2013}. A similar problem can be observed in other approaches, such as~\cite{Harville2001} and~\cite{Kolmogorov}.



\subsection{Recognizing people}
\label{sec:recognizing}
Multiple cues are also used to detect objects in image sequences, a survey about this subject with a Kinect (RGBD camera) can be found in~\cite{han2013enhanced}. 

Jun Liu~\textit{et al}. developed an automatic detection and tracking of people in cluttered and dynamic environments using a single RGBD camera. The original RGBD pixels are transformed to a Point Ensemble Image (PEI), they demonstrate that human detection and tracking in 3D space can be performed.They have some missing detections  mainly caused by depth data loss~\cite{liu2015detecting}.

Spinello and Arras created a new way to detect humans in images with the Histograms of Oriented Depth (HOD) detector~\cite{spinello2011people}. Authors developed a human detector using depth data instead of color data and fused the Histograms of Oriented Gradients (HOG) and HOD detectors together. HOG and HOD descriptors were classified then fused using a learned Support Vector Machines (SVM). 

Choi~\textit{et al}. combined five observation models: HOG, shape from depth data, front face detection, skin color, and motion detection. The image data was gathered with an RGBD camera and a Reversible-Jump Markov Chain Monte Carlo (RJ-MCMC) algorithm was used to detect people in a frame. Their results showed that the combination of different detection cues provided more reliable results and the advantages of different cues could overcome the disadvantages of other cues~\cite{choi2011detecting}.

Matt Davis and Ferat Sahin presented a novel  method  to  identify  humans by  combining  features  detected  in RGB, depth,  and  thermal  images.  They used  HOG  features  extracted from  the  three  image  types and created a multi-layer classifier  that did not overcome  a simple SVM  thermal  HOG  classifier~\cite{davis2016hog}.

Cristina Palmero~\textit{et al}. tried to address the problem of human body segmentation from multi-modal visual cues. They proposed a novel RGB-Depth-Thermal dataset. In order to classify regions of images as human or not humans they fused the features from each cue in a Gaussian Mixture Model (GMM). They classified patches of the image based on a random forest~\cite{palmero2016multi}.

\subsection{Aim of the paper}

Our aim is to create an unified model that mixes multiple information cues from the devices to segment foreground regions in image sequences. This regions are those ones that contains moving objects. We try to label those regions applying a state-of-the-art people detector. In this paper we present an experimental setup using a dataset that provides sequences in the RGBDT space, adapting our previous work~\cite{moya2017modeling} to include thermal information.

The paper is organized as follows. In Section~\ref{sec:Introduction}, we explain the context of the problem, the related work and our goal. In Section~\ref{sec:sequences}, we describe the proposed model to detect objects and the challenges of working with sequences in a multidimensional space. In Section~\ref{sec:experimentation}, we describe an experimental configuration of the proposed method with three different sequences and the preliminary results we obtained. Finally, we present the conclusions.

\section{Detecting objects in the RGBDT space}
\label{sec:sequences}

In order to select the regions with moving objects in an image sequence, we use a non-parametric algorithm that is capable to mix the color, depth and thermal information in a low-level way using the previous information as the reference to segment the current frame.

The scene modelling consists on a Kernel Density Estimation (KDE) process. Given the last $n$ observations of a pixel, denoted by $\mathbf{x}_i$ , $i = 1, \ldots, n$ in the d-dimensional observation space $ \mathbb{R}^d$, which enclose the sensor data values, it is possible to estimate the probability density function (pdf) of each pixel with respect to all previously observed values~\cite{elgammal2000non}.

\begin{equation} \label{eq:multydensity}
P(\mathbf{x}) = \frac{1}{n}|\textbf{H}|^{-\frac{1}{2}} \sum\limits_{i=1}^{n}K(\textbf{H}^{-\frac{1}{2}}( \mathbf{x}- \mathbf{x}_i)) \ ,
\end{equation} 

\noindent where $K$ is a multivariate kernel, satisfying $\int K(x)dx = 1$ and  $K(u)$ $\geq$ 0. \textbf{H} is the bandwidth matrix, which is a symmetric positive \textit{d$\times$d}-matrix.

The choice of the bandwidth matrix \textbf{H} is the single most important factor affecting the estimation accuracy because it controls the amount and orientation of smoothing induced~\cite{wand1994kernel}. 

Diagonal matrix bandwidth kernels allow different amounts of smoothing in each of the dimensions and are the most widespread due to computational reasons~\cite{dbandwith}. The most commonly used kernel density function is the Normal function, in our approach $N(0,\textbf{H})$ is selected

\[ \textbf{H} =\left( \begin{array}{cccc}
\sigma_1^2 & 0 & \cdots & 0 \\
0       & \sigma_2^2 & \cdots & 0 \\
\vdots  & \vdots     & \ddots & \vdots \\  
0 & 0 & \cdots& \sigma_d^2 \end{array} \right) \]

The final probability density function can be written as

\begin{equation} \label{eq:elgammalD}
P(\mathbf{x}) = \frac{1}{n} \sum\limits_{i=1}^{n} \prod\limits_{j=1}^{d}\frac{1}{\sqrt{2\pi\sigma_j^2}}e^{-\frac{1}{2}\frac{(x_{j} - x_{ij})^2}
{\sigma_j^2}} \ .
\end{equation}

\begin{figure}[!t]
    \centering
\begin{subfigure}[b]{0.48\columnwidth}
       {\includegraphics[width=\columnwidth]{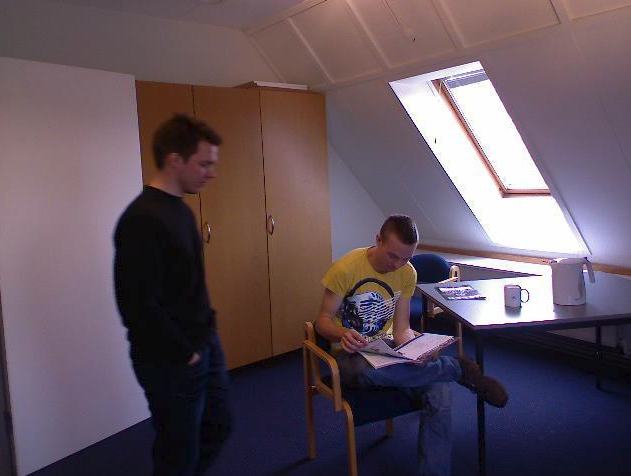}}
        \caption{RGB Image.} \label{fig_backsubs:a}
   	
    \end{subfigure}
     \begin{subfigure}[b]{0.48\columnwidth}
        
		{\includegraphics[width=\columnwidth]{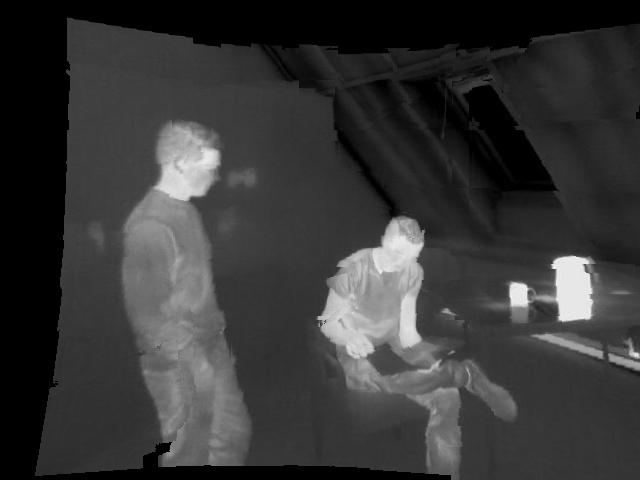}}
		 \caption{Thermal image.} \label{fig_backsubs:b}
	
    \end{subfigure}
    
    \begin{subfigure}[b]{0.48\columnwidth}
   		{\includegraphics[width=\columnwidth]{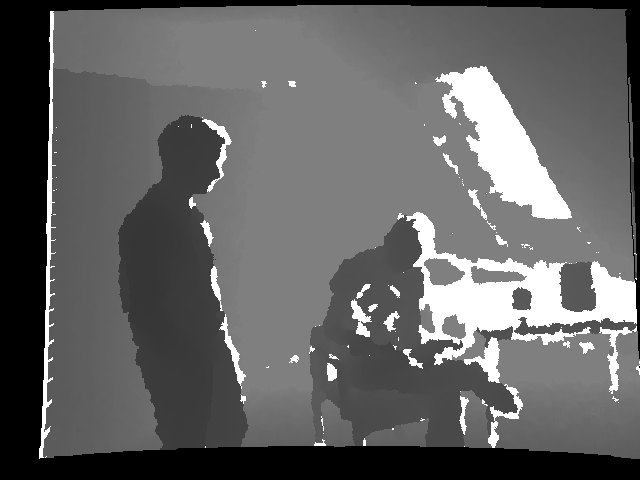}}
   		 \caption{Depth image.} \label{fig_backsubs:c} 
    \end{subfigure}
     \begin{subfigure}[b]{0.48\columnwidth}
		{\includegraphics[width=\columnwidth]{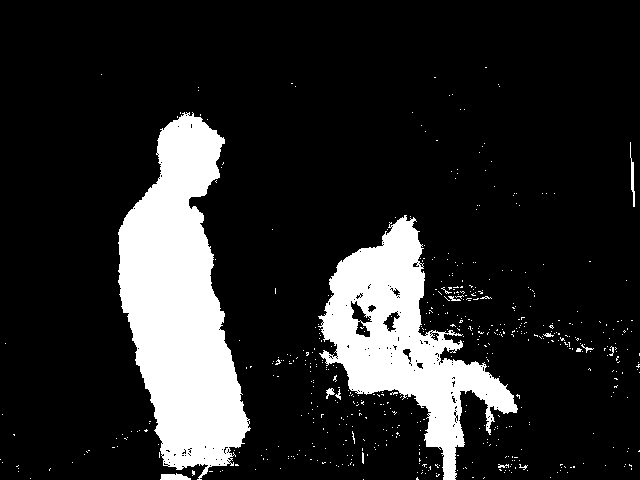}}
		 \caption{Segmentation result.} \label{fig_backsubs:d}
	
    \end{subfigure}
     
    \caption{Result of the modelling algorithm after 500 frames. Figures (a), (b) and (c) represent the input cues and (d) depicts the foreground mask.}
    	\label{fig_backsubs}
\end{figure}

\noindent Given this estimate at each pixel, a pixel is considered foreground if its probability is under a certain threshold, see Fig~\ref{fig_backsubs}.

From the scene modelling algorithm results we extract the Regions Of the Interest (ROI) of each frame. After applying a morphological opening operation, the ROI is defined as the bounding box of the remaining relevant blobs. Each region of interest should contain a invididual object instance. In our case, different objects may overlap in space, resulting in a bigger region that can contains more than one item.

\subsection{RGBDT issues}

Each type of information has its own particularities. We are not interested in creating a model for each cue, so we studied the characteristics of each one in order to include them properly into the unified model.

Usually color information is useful for suppressing shadows from detection by separating color information
from lightness information. To construct a robust algorithm that is independent of illumination variations,
we separated color information from luminance information using a non-luminance dependent color space. 
Chromaticity is the description of a color ignoring its luminance, and it can be described as a combination
of hue and saturation. Given the device's three color channels \textit{R, G, B}, the chromaticity coordinates
\textit{r, g} and \textit{b} are: $r = {R}/{(R+G+B)}$, $g = {G}/{(R+G+B)}$, $b = {B}/{(R+G+B)}$ where: $r+g+b = 1$~\cite{levine1985vision}. In our model we use two dimensions: $r$ and $g$.

The depth channel, D, has a significant amount of missing information from instances in which  the sensor is unable to estimate the depth at certain pixels.  For this purpose, we properly defined the Absent Depth Observations (ADO) to include them in the scene model by constructing a probabilistic model. Therefore, absent observations can be handled in a unified manner \cite{moya2017modeling}. 

Thermal channel, T, is similar to a grayscale color image. It seems that the
thermal cue can segment the human body more accurately, but it can include
some undesired reflections and illuminate warm objects
\cite{palmero2016multi}. It can be added directly as a new dimension to our
background model. As in indoor scenarios when people appears in the scene,
provokes an effect similar to switch-lighting, we decided to use a high
bandwith value with this channel in order to smooth this effect.

\begin{figure*}[!t]

    \begin{subfigure}[b]{\textwidth}
        {\includegraphics[width=0.245\textwidth]{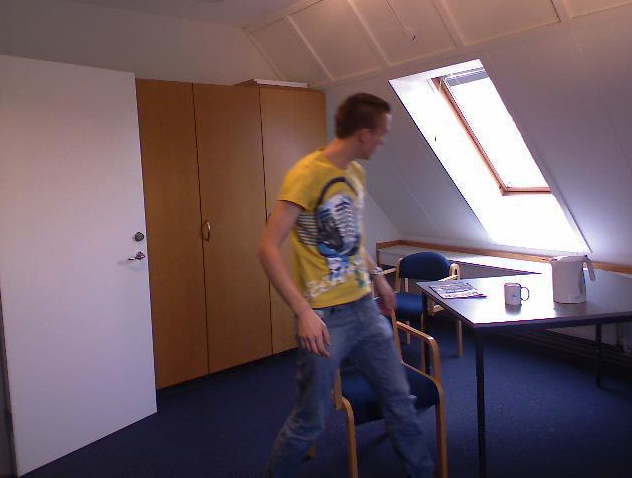}}
   		{\includegraphics[width=0.245\textwidth]{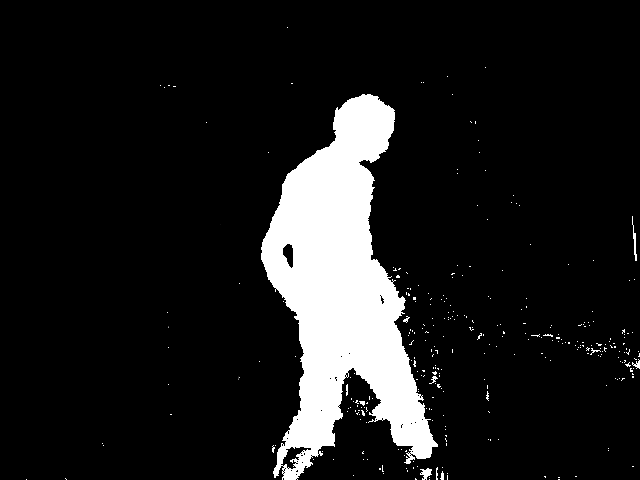}}
        {\includegraphics[width=0.245\textwidth]{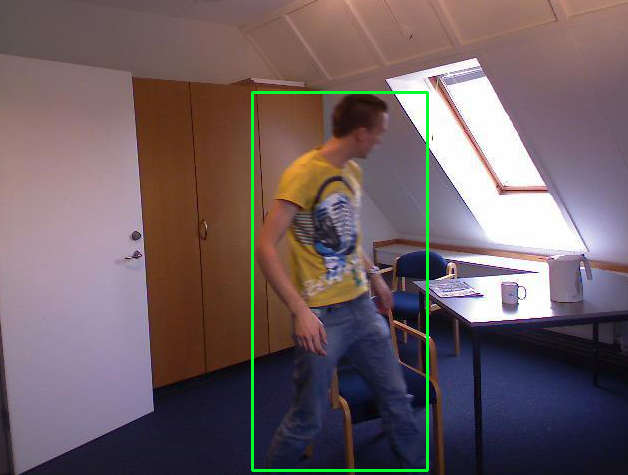}}
	    {\includegraphics[width=0.245\textwidth]{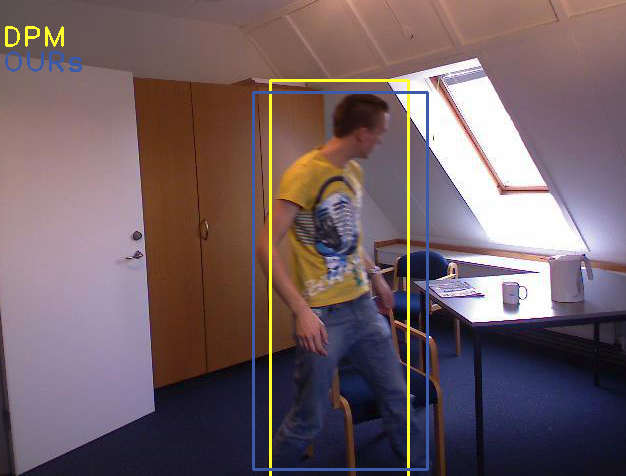}}
    \end{subfigure}
    \caption{Description of the whole process. First image corresponds to RGB frame. Second image depicts the foreground mask. Third image show the area labeled  as region of interest. Last image show the detection of a human (yellow box) by the algorithm described in~\cite{felzenszwalb2010object}.}
    	\label{fig_process}
\end{figure*}

\section{An initial experiment: recognizing people}
\label{sec:experimentation}

To test our approach we used a novel dataset. This dataset described in \cite{palmero2016multi}, features a total of 11,537 frames RGB-Depth-Thermal frames divided into three indoor scenes, of which 5724 were annotated. That contains up to three individuals who appear concurrently in three indoor scenarios, performing diverse actions that involve interaction with objects. The RGBDT data stream was recorded using a Microsoft Kinect for XBOX360, which captures the RGB and depth image streams, and an AXIS Q1922 thermal camera.

Scenes 1 and 2 were situated in a closed meeting room with little natural light to disturb the sense of depth, while scene 3 was situated in an area with wide windows and a substantial amount of sunlight. The human subjects were walking, reading, using their phones or interacting with each other.

\begin{figure}[!t]

    \begin{subfigure}[b]{\columnwidth}
        {\includegraphics[width=0.32\columnwidth]{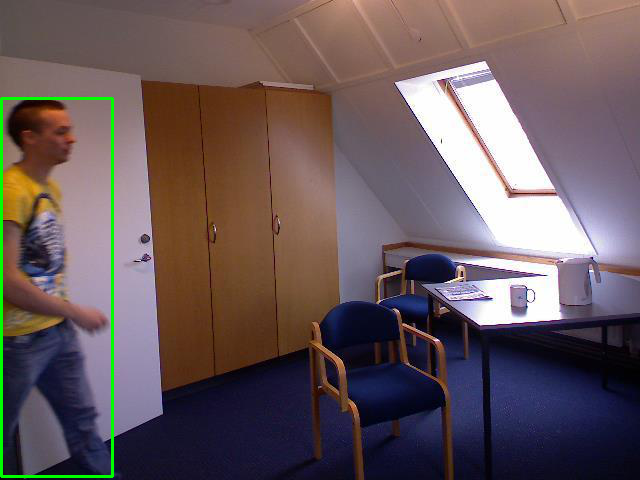}}
   		{\includegraphics[width=0.32\columnwidth]{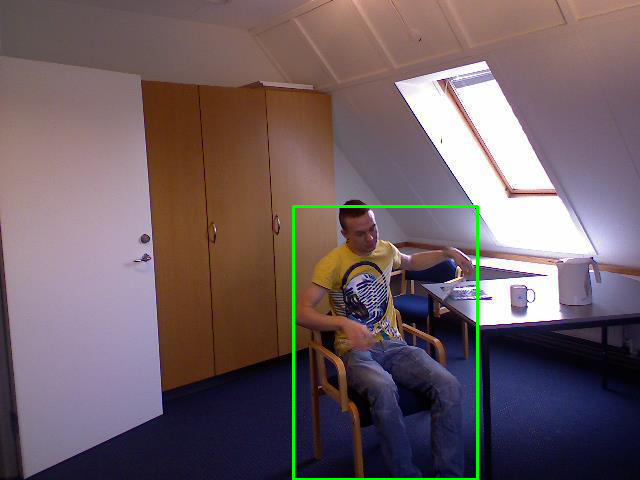}}
        {\includegraphics[width=0.32\columnwidth]{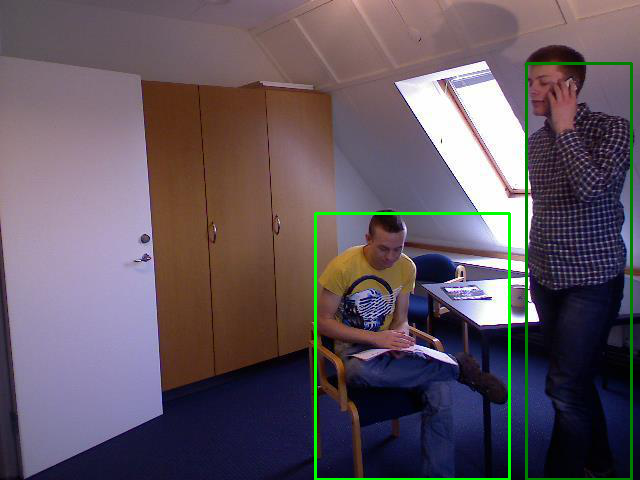}}
    \end{subfigure}
     \begin{subfigure}[b]{\columnwidth}
        {\includegraphics[width=0.32\columnwidth]{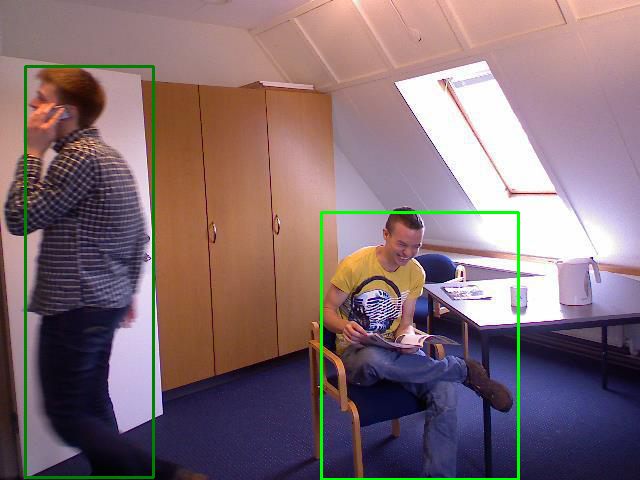}}
		{\includegraphics[width=0.32\columnwidth]{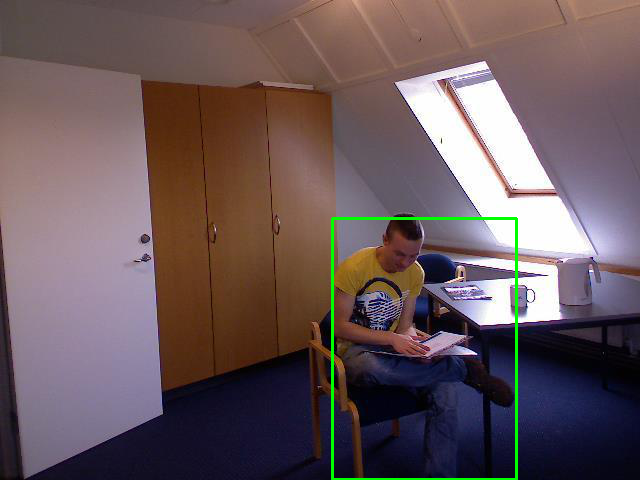}}
		{\includegraphics[width=0.32\columnwidth]{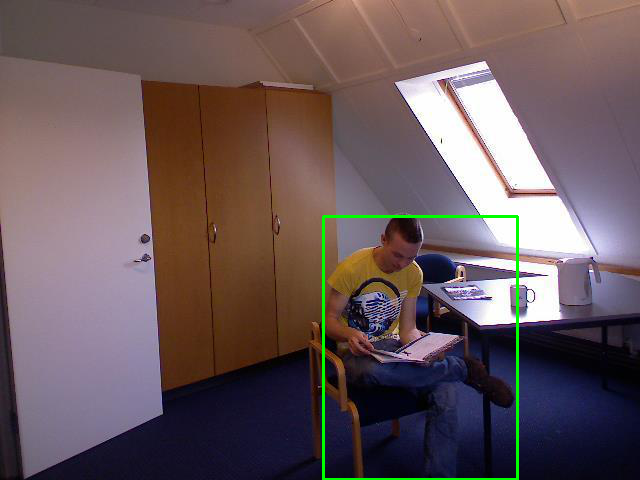} }
    \end{subfigure}
     \begin{subfigure}[b]{\columnwidth}
        {\includegraphics[width=0.32\columnwidth]{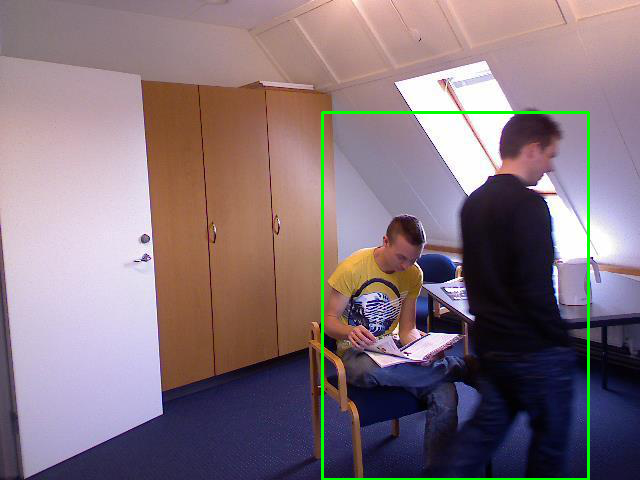}}
		{\includegraphics[width=0.32\columnwidth]{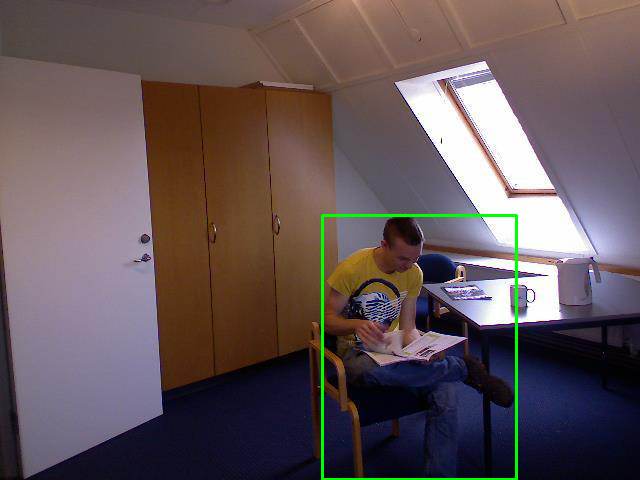}}
		{\includegraphics[width=0.32\columnwidth]{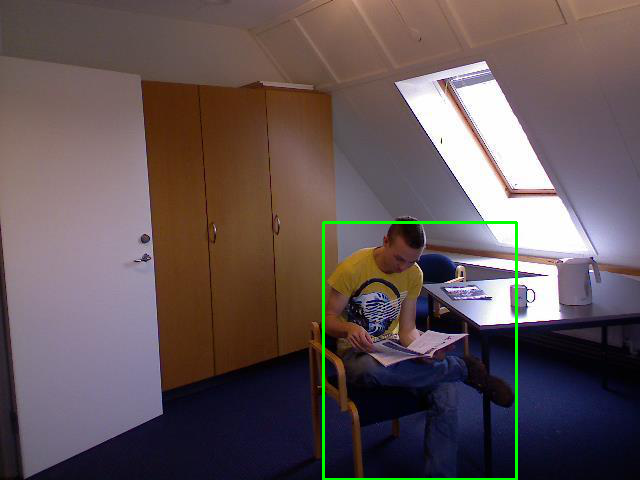}}
    \end{subfigure}
    \caption{Results of our algorithm applied to the second sequence of the dataset.}
    	\label{fig_TestII}
\end{figure}

\begin{figure}[!t]

    \begin{subfigure}[b]{\columnwidth}
        {\includegraphics[width=0.32\columnwidth]{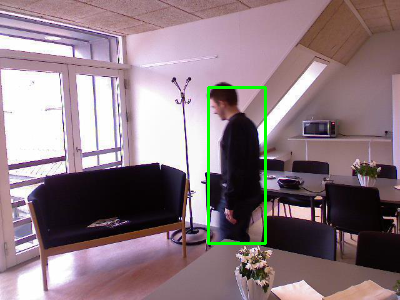}}
   		{\includegraphics[width=0.32\columnwidth]{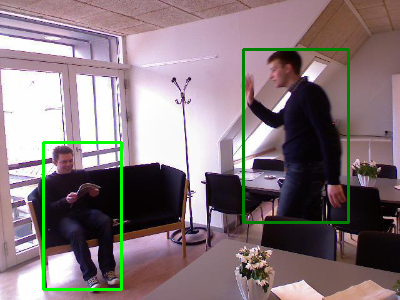}}
        {\includegraphics[width=0.32\columnwidth]{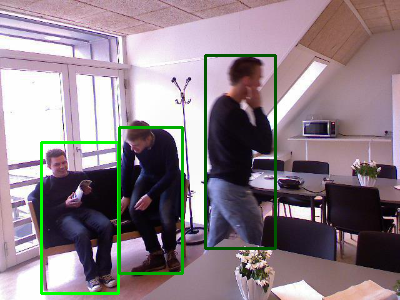}}
    \end{subfigure}
     \begin{subfigure}[b]{\columnwidth}
        {\includegraphics[width=0.32\columnwidth]{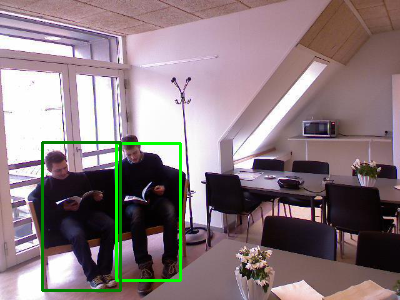}}
	    {\includegraphics[width=0.32\columnwidth]{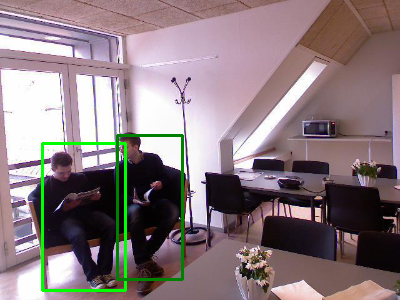}}
		{\includegraphics[width=0.32\columnwidth]{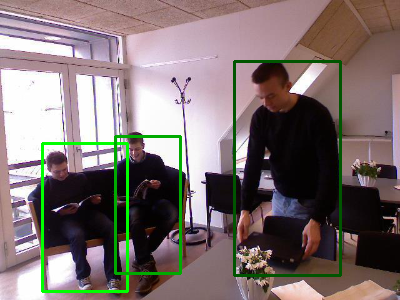}}
    \end{subfigure}
     \begin{subfigure}[b]{\columnwidth}
        {\includegraphics[width=0.32\columnwidth]{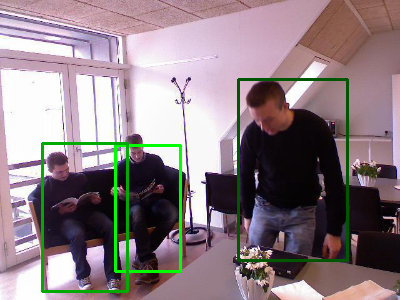}}
		{\includegraphics[width=0.32\columnwidth]{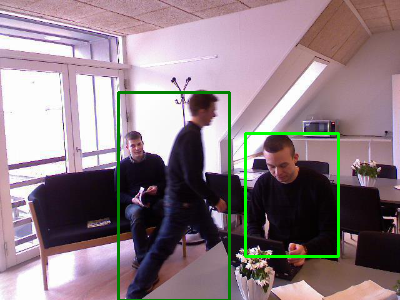}}
		{\includegraphics[width=0.32\columnwidth]{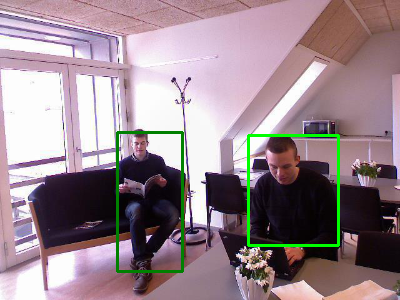}}
    \end{subfigure}
    \caption{Results of our algorithm applied to the third sequence of the dataset.}
    	\label{fig_TestIII}
\end{figure}

Results of detecting objects algorithm  are depicted in Fig.~\ref{fig_TestII} and Fig.~\ref{fig_TestIII} from Scenes 2 and 3. We can observe that after applying the scene modelling process we obtained the ROIs with the moving objects.

Next step was to label the selected regions that corresponds to human instances. A first approach to check the viability of the proposed solution to recognize people in image sequences was to apply the state-of-the-art human detector described in~\cite{felzenszwalb2010object} to the ROIs of each frame,. See the whole process in Fig.~\ref{fig_process}. 

Only one out of fifteen people instances that appear in the three sequences weren't labeled due to its small time in the scene and its strange pose.



\section{Conclusions}

In this article we presented an approach to label regions that contain moving objects in image sequences by applying a probabilistic non-parametric model that mixes multiple information cues. We prepared an experimental setup using a dataset that provides sequences in the RGBDT space to show the importance of using previous information in order to obtain an accurate segmentation result. We adapted our previous model~\cite{moya2017modeling} to add a new cue: thermal information. We labeled people in the regions with moving objects using a state-of-the art algorithm. All but one of our detected regions is at least labeled as human in one frame.

{\small
\bibliographystyle{ieee}
\bibliography{egbib}
}

\end{document}